%% file: iclr2027_conference.tex
\pdfoutput=1

\documentclass{article} 
\usepackage{iclr2027_conference,times}

\input{math_commands.tex}

\usepackage{hyperref}
\usepackage{url}

\usepackage{graphicx}
\usepackage{multirow}
\usepackage{enumitem}
\usepackage{amsmath}
\usepackage{cleveref}
\usepackage{makecell}
\usepackage{caption}
\usepackage{booktabs}

\usepackage{color}
\usepackage{CJKutf8}
\usepackage{lineno}

\hypersetup{
    colorlinks=true,
    linkcolor=red,
    urlcolor=cyan,
    citecolor=cyan,
}

\newcommand\textblue{\textcolor[RGB]{3, 111, 193}}
\newcommand\textorange{\textcolor[RGB]{255, 153, 0}}

\newcommand{\resultbest}[1]{\textorange{\textbf{#1}}}
\newcommand{\resultsecond}[1]{\textblue{\underline{#1}}}

\title{We Have Been Interpreting Diffusion\\Transformer Backbone Behaviors Wrong}

\author{\parbox{14cm}{
    \centering
    \large{
        Benyuan Meng$^{1\dagger}$ \quad
        Longtao Huang$^{2\dagger}$ \quad
        Yang Liu$^{1,4}$ \quad
        Liang Lin$^{1,3,4}$
    }
    \\ \vspace{2pt}
    \small{$^\dagger$Equal contribution}
    \\ \vspace{6pt}
    \normalsize{
        \textnormal{$^1$Sun Yat-sen University}\quad
        \textnormal{$^2$Alibaba Group}\quad
        \textnormal{$^3$Peng Cheng Laboratory}\quad
        \textnormal{$^4$X-Era AI Lab}
    }
    \\ \vspace{4pt}
    \texttt{mengby6@mail2.sysu.edu.cn}\quad
    \texttt{kaiyang.hlt@alibaba-inc.com}\quad
    \texttt{liuy856@mail.sysu.edu.cn}\quad
    \texttt{linliang@ieee.org}
}}

\iclrfinalcopy 
\begin{document}

\maketitle

\begin{abstract}
During a network forward call, diffusion backbones naturally produce many intermediate activations.
If we evaluate each activation against an identical downstream benchmarking protocol, we can plot an inverted-U-shaped curve as the depth profile: the activation's performance goes up first and then down as we go into deeper layers.
This profile suggests there is a semantically meaningful signal within the backbone, and it is changing.
This is common knowledge in the diffusion model community, which inspires many applications including the widely-adopted REPA.
However, the common interpretation of this depth profile may be inaccurate and incomplete:
(i) The cause of this curve has long been treated as signal granularity changing from coarse to fine, but there has been little evidence in diffusion transformers.
(ii) The community has been mainly studying this phenomenon in a static way, without investigating how it changes across different training setups.
In this work, we show a different mechanism picture for the profile, where the prediction target (\textit{e.g.}, $\epsilon$, $u$, $x_0$) sits at the center.
The semantically meaningful signal is only an intermediate by-product for predicting the target, while the prediction target is the real goal for the backbone.
Furthermore, the choice of prediction targets as part of the training setup can predictably affect the curve's shape.
With this updated interpretation, many applications of this depth profile can be better grounded and predicted.
As an example, we show how to predict the near-optimal placement of REPA based on the prediction target of the training setup.
The code is \href{https://github.com/Darkbblue/dit-compare}{here}.
\end{abstract}

\vspace{0.5em} 
\begin{center}
    \centering
    \includegraphics[width=0.95\linewidth]{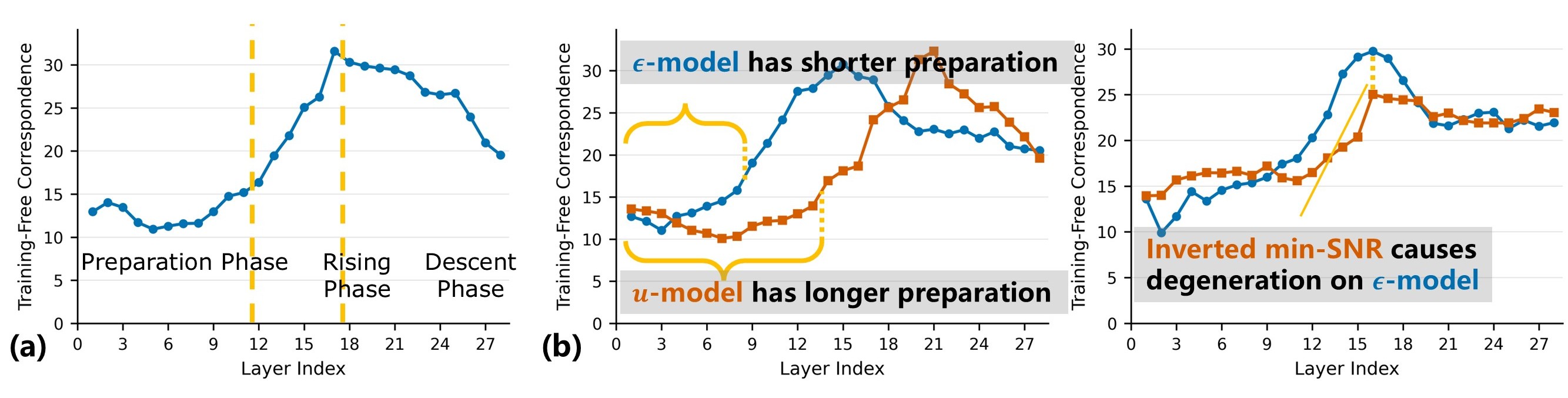} 
    \captionof{figure}{(a) The inverted-U-shaped curve, with layer index as x-axis and correspondence performance (or other downstream task performance) as y-axis, is common knowledge in the diffusion community.
    (b) Two representative \textit{new} backbone behaviors among our findings, where \textit{the training setups} determine the depth profile. Left: prediction target determines the length of the preparation phase. Right: prediction target can lead to degeneration under certain settings, indicated by the significantly lower peak height.}
    \label{fig:intro}
\end{center}
\vspace{0.5em} 

\input{sections/1_intro}
\input{sections/2_related}
\input{sections/3_preliminary}
\input{sections/4_setup}
\input{sections/5_content}
\input{sections/6_example}
\input{sections/7_conclusion}

\subsection*{AI use statement}

(This section is \textbf{required} and does not count toward the page limit.)

In this work, we used generative AI tools to:
(i) Propose the necessary components for the alignment training framework, though the idea of alignment itself is proposed by humans.
(ii) Code, with necessary human reviewing.
(iii) Help interpret probing results and design further probes to dig deeper into the backbones, though quite often human intelligence is still required where LLMs fail. For instance, 2 in 4 of the major breakthroughs during the research process is mainly made by humans while the other 2 by LLMs.
(iv) Provide feedback for the writing, though we do not directly use LLMs to refine writing, because of personal taste. Also, we notice that the LLM feedback often focuses too much on details that contribute little to our main claim/contribution, requiring generally less meaningful ablation experiments to be added.
We have not used generative AI tools for finding motivations and ideas / help develop theoretical models or conceptual frameworks,
and generate synthetic data sets / formulate mathematical claims / provide critical ingredients for proving mathematical claims / assist in the writing of proofs / propose or refine hypotheses /assist with translation / clean and reformat dataset / support qualitative and thematic data analysis are not applicable to this work.

Additionally, we used generative AI tools for:
(i) Create matplotlib codes to render the plots (but the illustrative figures are drawn by humans). This is reliable by nature.
(ii) Summarize or analyse existing literature (using the GPT-Researcher harness to enhance reliability, along with necessary human reviewing).
(iii) Format references. For example, when directly copying the DBLP bibtex citations, the conference titles are often not directly usable, or the page number may be missing for certain conferences or ArXiV papers.
We use an agentic LLM rather than a pure-text-based LLM and develop a skill to enhance reliability.

We have reviewed all AI-assisted work or use other means to enhance reliability, as described immediately after each usage.
We take responsibility for the final content of this work,
including text, claims or artifacts produced with the aid of generative AI.

\subsection*{Ethics statement}
This is a general diffusion model study and does not seem to relate to ethical issues.

\subsection*{Reproducibility statement}

We plan to release our codes after acceptance, but not during reviewing.
When releasing, we will include both training codes and probing codes, along with the command we used.


\bibliography{iclr2027_conference}
\bibliographystyle{iclr2027_conference}

\newpage
\appendix
\input{sections/appendix}

\end{document}

%% file: math_commands.tex
\usepackage{amsmath,amsfonts,bm}

\def\eqref#1{equation~\ref{#1}}

\def\1{\bm{1}}

\DeclareMathAlphabet{\mathsfit}{\encodingdefault}{\sfdefault}{m}{sl}
\SetMathAlphabet{\mathsfit}{bold}{\encodingdefault}{\sfdefault}{bx}{n}



%% file: sections/1_intro.tex
\section{Introduction}

Diffusion models are a trending type of generative model, mainly used for, but not restricted to, image generation, consisting of a scheduler and a backbone network~\citep{ho2020denoising, dhariwal2021guideddiffusion, rombach2022high, podell2023sdxl, wu2025qwenimagetechnicalreport}.
As a neural network, the diffusion backbone produces many intermediate activations during a forward call, which can be denoted as diffusion features~\citep{baranchuk2021label, xu2023open, zhao2023unleashing, tang2023emergent, zhang2023tale, luo2023dhf, my-SDXL, ma, DBLP:conf/iclr/Chen0XH25}.
Evaluating diffusion features' performance on downstream vision benchmarks can be used as a \textit{probe} to understand diffusion backbone behaviors~\citep{REPA}.
One important instance of these behaviors is the depth profile: the performance first goes up with layer depth and then down, forming an inverted U-shape, as shown in \Cref{fig:intro}(a).
This suggests that there is a semantically meaningful signal within the backbone, and it changes across layers.
To name its importance, REPA~\citep{REPA}, optimal layer selection~\citep{zhou2023hyperspectral, luo2023dhf}, and post-hoc normalization~\citep{ma}, are all directly or indirectly based on the interpretation of the depth profile.

However, the current interpretation of the depth profile is potentially inaccurate and incomplete.
On one hand, the popular opinion is that the semantic signal gradually changes its granularity from coarse to fine, and the peak is at the exact position where the granularity best suits the benchmarking task~\citep{baranchuk2021label, luo2023dhf, REPA}.
However, this is not supported by strong evidence after the community moves to diffusion transformers (DiT), where there are no resolution changes to force such a behavior as in U-Nets.
On the other hand, previous interpretations are mostly static, focusing on one model and overlooking the potential influence of the training setups on the depth profile.
Therefore, the current application of the depth profile property relies on a foundation that is still partially black-box.
In this work, we provide a more complete and likely more accurate picture, where the \textit{prediction target} sits at the center.
Here, prediction target means what type of signal the backbone actually predicts:
not all diffusion backbones predict noise $\epsilon$, as they can also predict $u$\footnote{The so-called $v$ in DDPM literature~\citep{DBLP:conf/iclr/SalimansH22} and flow matching literature~\citep{flow-matching} is different, and we define $u = \epsilon - x_0$ in place of the ambiguous $v$.} or $x_0$ instead, which can all be utilized for denoising with proper formulas~\citep{jit}.

We reveal that the whole depth profile can be divided into three phases, as shown in \Cref{fig:intro}(a), and that the real goal of the backbone is to predict the target, with a semantically meaningful inner representation as only a by-product.
(i) \textbf{Preparation phase}:
The semantic signal helpful for the downstream task has \textit{not} started to form. Instead, the model extracts $\epsilon$ and/or $x_0$ from the residual\footnote{For a residual connection $x\leftarrow x+\text{Module}(x)$, $x$ is residual and Module($x$) is increment.} and stores them back into subspaces for later use.
(ii) \textbf{Rising phase}:
A highly abstract and high-level representation $x_s$ of the clean image gradually forms, peaking at a certain layer, which is exactly the semantic signal for the downstream task.
(iii) \textbf{Descent phase}:
The model combines the semantic $x_s$ with $\epsilon$ and/or the input-space $x_0$ to encode the prediction target.
The semantic signal is therefore still within the residual stream, with a new signal corresponding to the prediction target being added to the residual stream.
Hence, at least one major cause for the curve going down is the new signal overrunning the semantic signal, beside the granularity change.

Furthermore, we find that the training setups can affect, or even \textit{determine}, the depth profile in a way that the prediction targets play a vital role, as shown in \Cref{fig:intro}(b).
One example corollary from the findings above is that the length of the preparation phase is determined by the prediction target, where a more \textit{hybrid} target leads to longer preparation than a \textit{purer} target.
Here, we treat targets like $x_0$ or $\epsilon$ as pure, and targets like $u=\epsilon-x_0$ as hybrid, as it mixes the two pure targets into one.
Another example is that highly pure targets can lead to \textit{degeneration} when paired with certain loss weighting.
Both corollaries align well with the actual observations.

Lastly, to exemplify the practical implication of our findings, we show how to predict the near-optimal placement for REPA~\citep{REPA} based on our findings.
We even make an aggressive prediction that the REPA alignment layer should be gradually moved to earlier layers as the training goes on, which is also validated.
These demonstrations serve as falsification tests of the mechanism interpretation rather than as method contributions.
Our contribution is as follows:
\begin{itemize}
    \item We propose and validate a more complete and likely more accurate mechanism interpretation for the diffusion backbone depth profile, with prediction targets playing a vital role.
    \item We reveal how training setups, especially prediction targets, determine the depth profile.
    \item Practically, our interpretation makes certain practice ambiguities less black-box and more predictable, alleviating the need for exhaustive search or intuition.
\end{itemize}

%% file: sections/2_related.tex
\section{Related Work}
\textbf{Diffusion Feature.}
Diffusion feature has become an acknowledged research area through years of advancements.
Early on, the concept of using diffusion backbones as feature extractors was validated by pioneer studies such as \citet{baranchuk2021label, xu2023open, zhao2023unleashing, tang2023emergent}.
In such early studies, the depth profile of diffusion backbones is already observed and utilized to select the near-optimal layer to boost downstream performance.
However, this selection process relies on heavy ablation or intuition, so researchers have developed some techniques for the selection, such as an automatic layer selector~\citep{tang2023emergent, zhou2023hyperspectral}.
This shows that the depth profile still remains partially black-box in nature.
Additionally, whilst most diffusion feature studies focus on U-Net-based backbones, new findings on DiT backbones are emerging~\citep{ma, DBLP:journals/corr/abs-2502-04320}.

\textbf{Probing Diffusion Models.}
There are different ways to probe a diffusion model and gain a better understanding of its behaviors and properties.
(i) Diffusion feature, as introduced before, can serve as a probe.
(ii) Another common probe is specifically the attention maps within cross-attention layers, calculated via the matrix multiplication of $Q$ and $K$~\citep{daam}.
This could help interpret how a diffusion backbone aligns textual semantics to pixels~\citep{DBLP:conf/cvpr/MengXWCHH26}.
(iii) Sparse Auto Encoder (SAE) is a recently emerging tool for diffusion model probing~\citep{DBLP:conf/nips/SurkovWMTDWGB25, DBLP:conf/nips/TinazFS25}.
(iv) REPA~\citep{REPA} uses DINOv2 features as an external alignment target and measures how well each diffusion feature aligns with them.

Our work lies in the intersection of the two aforementioned directions: we use diffusion feature on downstream vision tasks as a way to probe diffusion models.
In this study, we do \textit{not} aim to improve diffusion feature performance, but only to better understand model behavior and mechanisms.

%% file: sections/3_preliminary.tex
\section{Preliminary \& Notation}
Different from the common formalization of DDPM~\citep{ho2020denoising, song2020score} or flow matching~\citep{flow-matching}, we will use a more general way to describe the diffusion theoretical framework.
A diffusion model consists of a scheduler and a backbone model (Net).
The scheduler is used to produce training data at training time and apply the model prediction to the data point at inference time.
To construct one data sample, we need an input image $x_0$, a sampled random noise $\epsilon$, a timestep $t$, and optionally a condition $c$.
In this manuscript, we use $t=0$ to denote the clean end and $t=1$ for the noisy end.
Note that the input image is commonly compressed into VAE space~\citep{rombach2022high}, but we do not explicitly use another notation such as $z_0$ for this, for simplicity.
With $x_0$ and $\epsilon$, the scheduler constructs an $x_t = \alpha(t)x_0 + \beta(t)\epsilon$ as linear combination.
Next, the scheduler also constructs a ground truth $\text{GT} = \alpha(t)'x_0+\beta(t)'\epsilon$, where GT and $x_t$ can use different linear combination recipes.
The formulation of the ground truth is what we call \textit{prediction target}, as it is exactly what the backbone learns to predict.
Thus, we obtain a data point $(x_t, c, t, \text{GT})$.
Finally, the training objective is $ L = ||\text{Net}(x_t, c, t)-\text{GT}||^2$.
When generating, the scheduler iteratively calls the backbone at different $t$.
When extracting diffusion features, however, we mimic the \textit{training} procedure, meaning that we only feed $x_t$ into the backbone once at a selected $t$.

%% file: sections/4_setup.tex
\section{Probing Setups}
\subsection{Aligned Training Framework}
\subsubsection{Alignment}
\begin{figure}[t]
    \centering
    \includegraphics[width=\linewidth]{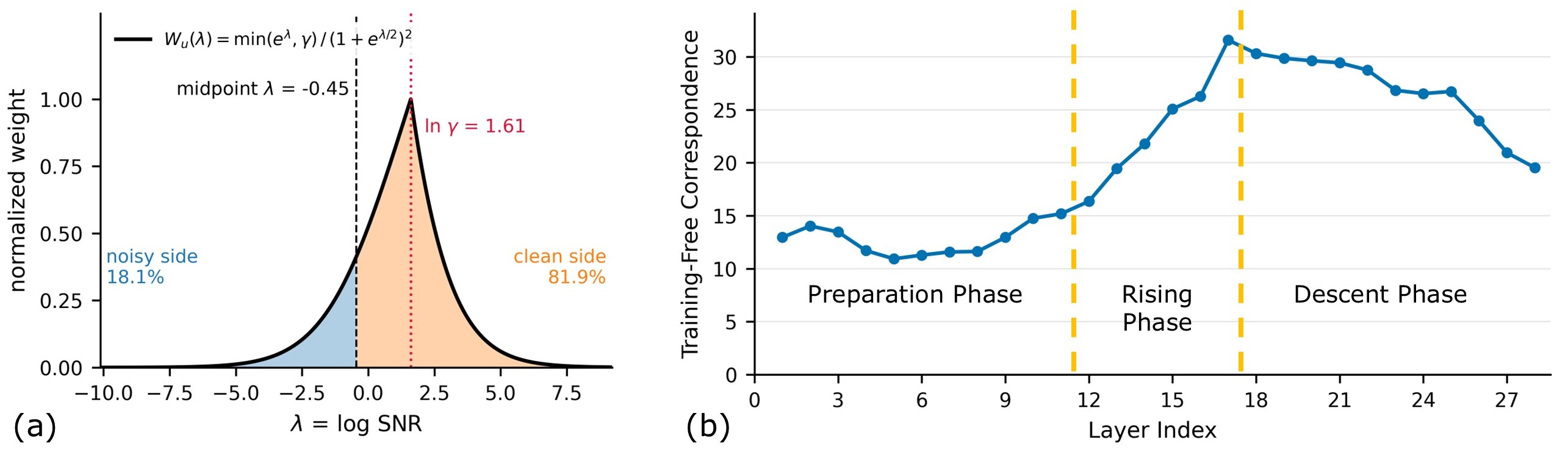}
    \caption{(a) A visualization of the default min-SNR weighting ($\gamma=5$) in the $u$-space.
    (b) A typical depth profile of diffusion backbones. The curve is obtained from FM+$u$ model with clean weighting.}
    \label{fig:min-snr}
\end{figure}

The motivation for the aligned training framework is to align most factors and only expose certain ones of interest within the training setups.
We implement the following alignment designs:

\textbf{Cross-Target Loss Weighting Alignment}.
Different prediction targets can be transformed into each other, but this process produces certain loss weighting terms that are functions of SNR (signal-to-noise ratio)~\citep{min-snr}.
For example, $||\epsilon-\hat\epsilon||^2=\text{SNR}(t)||x_0-\hat{x_0}||^2$, with the weighting term being $f(\text{SNR}) = \text{SNR}$.
To align this difference, we cast all prediction targets into $x_0$-space and cancel the derived coefficient.
For example, we can multiply $\epsilon$ loss with $\frac{1}{\text{SNR}}$.
Furthermore, for more stable training, we apply min-SNR weighting~\citep{min-snr} after the alignment, which emphasizes cleaner samples.
A visualization of the weighting curve in the ($u=\epsilon-x_0$)-space can be found in \Cref{fig:min-snr}(a).

\textbf{SNR Sampling}.
Conventionally, diffusion training setups sample timestep $t$ uniformly.
However, considering the different contours of sampling paths, uniform $t$ actually leads to non-uniform SNR, which can effectively be seen as implicit loss weighting~\citep{logsnr-sampling}.
Hence, we sample uniformly in the $log$SNR space instead, and the sampling space is [-10.12, 9.21] (which is why the midpoint in \Cref{fig:min-snr}(a) is -0.45).

\textbf{Training Steps}.
We train all models for 200k steps, which strikes a balance between time cost and model convergence, unless otherwise stated.

\textbf{Backbone Architecture}.
We use a backbone derived from DiT-XL/2~\citep{Peebles23dit} and SiT-XL/2~\citep{sit} as the main backbone, unless otherwise stated.
Specifically, the timestep embedding layer is modified such that both DDPM and flow matching paths use the same embedding protocol.
In this way, the backbone architecture is entirely aligned, only exposing the difference in training setups.
The DiT backbone, compared to U-Net backbones, is cleaner with fewer confounding factors such as resolution changes, making it suitable for model probing studies.

\textbf{Probe-Time SNR Alignment}.
It is a known property that the depth profile changes with the timestep for probing~\citep{baranchuk2021label, zhou2023hyperspectral, luo2023dhf}.
Furthermore, the timestep definition differs between DDPM and flow matching.
Hence, we do not probe models at matched timesteps, but matched SNR instead.

\subsubsection{Exposed Factors}
\textbf{Prediction Targets}.
We include practical targets for diffusion training: $\epsilon$, $u$, and $x_0$.
Furthermore, we design a diagnostic target $a(\theta)=cos\theta \epsilon - sin\theta x_0$, where $\theta$ is a hyperparameter fixed for the entire training process.
This allows testing finer-grained prediction targets.

\textbf{SNR Weighting}.
The base min-SNR weighting emphasizes the cleaner samples, so we denote it as the \textit{clean} weighting.
We can \textit{replace} the base min-SNR weighting with two alternative implementations: (i) a mirrored distribution to the curve shown in \Cref{fig:min-snr}(a) (the \textit{noisy} weighting), and (ii) a curve that is completely symmetric in the $log$SNR sampling space (the \textit{neutral} weighting).

\textbf{Sampling Path}.
We test the classic DDPM path (denoted as variance-preserving, VP) and the straight flow matching path (denoted as FM).

\subsection{Probes}
\textbf{Downstream Task Performance}.
(i) Following DIFT~\citep{tang2023emergent}, we use \textbf{training-free semantic correspondence} on SPair-71k dataset~\citep{min2019spair} as the first probe in our study, detecting semantically rich representations.
As a training-free probe, it can only detect signals that are directly interpretable without decoding.
(ii) To complement its capability, we also optionally \textbf{add one linear convolutional layer} to the probe, giving it \textit{linear} decoding capability.
Note that only the decoding capability changes, but the signal granularity requirement, which is determined by the semantic correspondence task, does not change.
(iii) We can also use \textbf{post-hoc transforms} on the features before correspondence computation, including whitening~\citep{whitening} (roughly similar to normalization) and transplantation~\citep{transplantation} (a free-form linear transform that projects a feature into the linear space of another given feature).
All these probes are measured using $log$SNR=5.88 unless otherwise stated.

\textbf{Other Probes}.
(i) \textbf{Linear regression} ($R^2$) predicts a target signal from an input signal via linear training or closed-form solutions~\citep{DBLP:conf/icml/HuhC0I24, DBLP:conf/iclr/AlainB17}.
(ii) \textbf{Separability probe}~\citep{separability} removes a direction from a given feature, but with the restriction of minimizing the impact on other given features that should be protected.
The cleaner the target direction can be removed under this restriction (indicated by \textit{lower} remaining $R^2$), the more separable these features are.
(iii) \textbf{Geometry probe} detects the relative norm and the angle between two features.
These probes are measured under $log$SNR=0 unless otherwise stated.

\textbf{Causal Intervention}.
We can manipulate certain features during inference and observe the influence on the outcome, including MSE loss and the main probe's curve.
Such manipulation can be done in two ways:
(i) Simply removing a certain direction from a feature;
(ii) Replacing a certain direction within one feature with a random noise of the same mean and variance as the replaced direction.

%% file: sections/5_content.tex
\section{Mechanism Findings}
As shown in \Cref{fig:min-snr}(b), the common depth profile of diffusion backbones is in an inverted U shape.
We divide this shape into three phases: preparation phase, rising phase, and descent phase.
\textit{The overall picture} is that the actual task for the diffusion backbone is to regress the prediction target $\text{GT} = \alpha(t)'x_0+\beta(t)'\epsilon$, and the forming of an inner semantic representation of the clean image is only a \textit{by-product}.
As shown in \Cref{fig:geometry-a45}(a):
(i) During the preparation phase, the model extracts $x_0$ and $\epsilon$ from the residual and stores them back into subspaces within the main residual.
(ii) During the rising phase, a semantic, abstract inner representation of the clean image, denoted as $x_s$ (``s'' for semantic), is formed.
(iii) During the descent phase, the model combines the semantic $x_s$ and the VAE-space signal $x_0, \epsilon$ to encode the prediction target and write it back into the residual, waiting to be decoded by the final linear layer.

\begin{figure}[t]
    \centering
    \includegraphics[width=\linewidth]{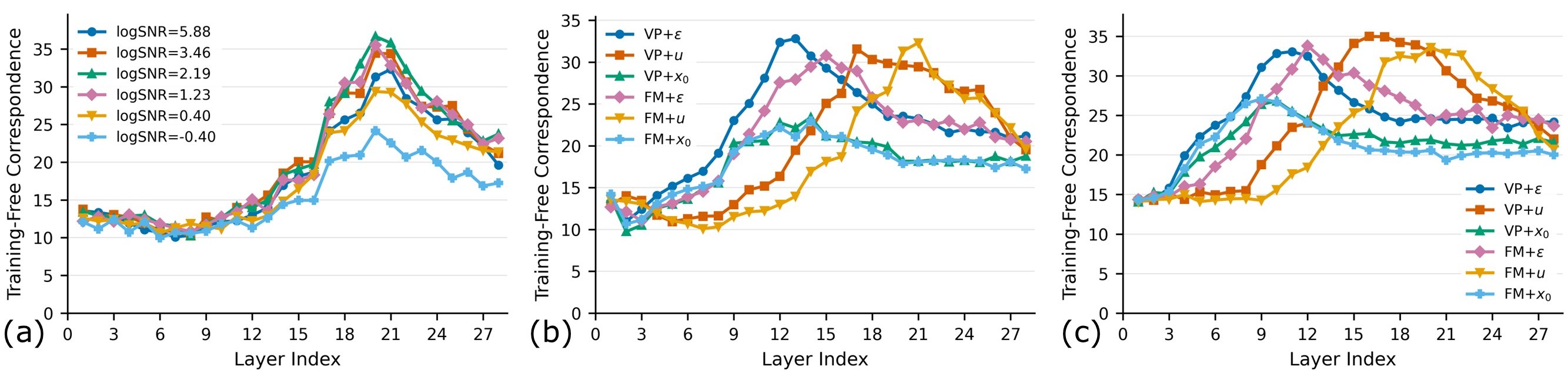}
    \caption{(a) Probing the same FM+$u$ model at different SNR yields highly similar depth profiles.
    (b) Basic training-free correspondence probe results on different sampling paths.
    (c) Whitening correspondence probe results on different sampling paths.}
    \label{fig:path+snr}
\end{figure}

Before going into details, we find that the depth profile is mainly static with respect to the inference-time SNR choice (except for the height of the peak), almost solely determined by the training dynamics (\Cref{fig:path+snr}(a)).
The peak height difference is related to the timestep profile, where a noisier timestep yields coarser semantics and a cleaner timestep yields finer-grained features~\citep{tang2023emergent}.
In addition, we find that the difference between sampling paths is relatively slight.
Considering the difference caused by prediction targets and loss weighting is already a lot, we will neglect the difference between sampling paths for now and leave it for future work (\Cref{fig:path+snr}(b,c)).

\subsection{Preparation Phase}
Based on the overall picture, the preparation phase prepares $x_0$ and $\epsilon$ by storing them into subspaces, and the prepared signals are mainly consumed \textit{not} in the rising phase but in the descent phase.

\begin{figure}[t]
    \centering
    \includegraphics[width=.95\linewidth]{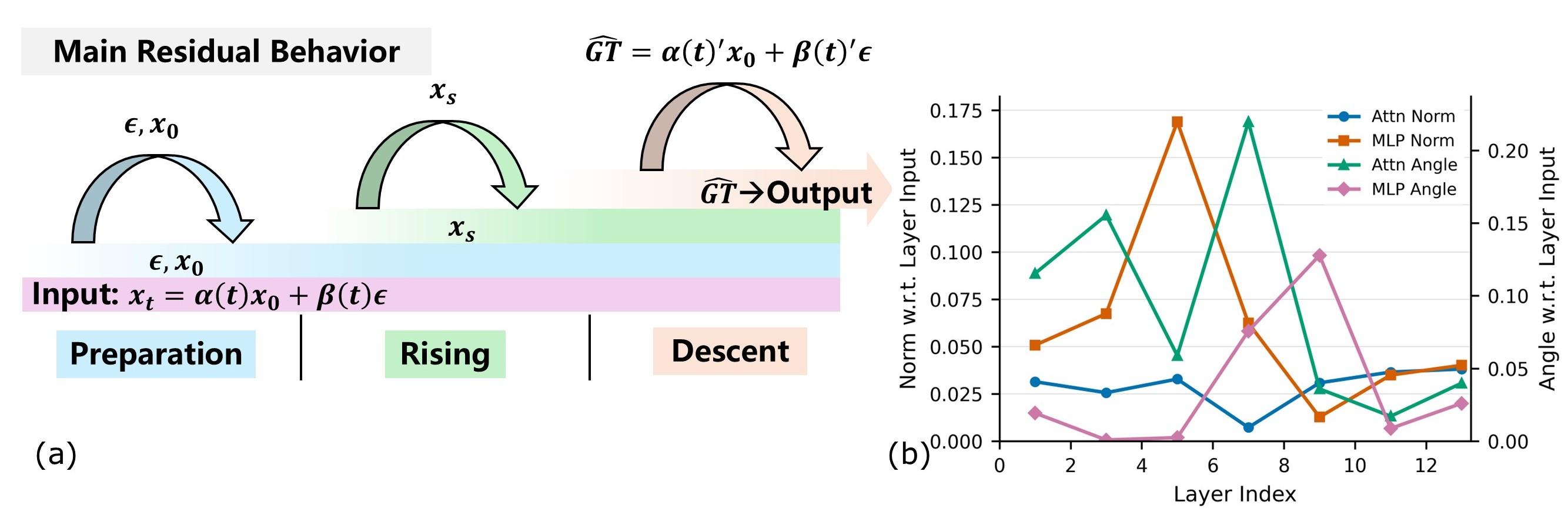}
    \caption{(a) An illustration of what each phase writes into the main residual and \textit{remains there afterward}.
    (b) Geometry probe results of FM+$a45$ model, angle measured in cosine. Each feature is measured relative to the layer input.}
    \label{fig:geometry-a45}
\end{figure}

First, we locate how the preparation is done.
Within each layer of a DiT backbone, there are an attention layer and an MLP layer.
With the geometry probe results in \Cref{fig:geometry-a45}(b), attention increments align better with the residual's direction but have smaller norms, while MLP increments are more orthogonal to the residual but have larger norms.
Then, we selectively zero out the attention increments or the MLP increments, finding that zeroing out the MLP increments leads to a more significant performance drop (MSE loss=0.3863 for attention and MSE loss=0.4176 for MLP).
This shows that MLP increments are more important to the preparation behavior.

\begin{figure}[t]
    \centering
    \includegraphics[width=.95\linewidth]{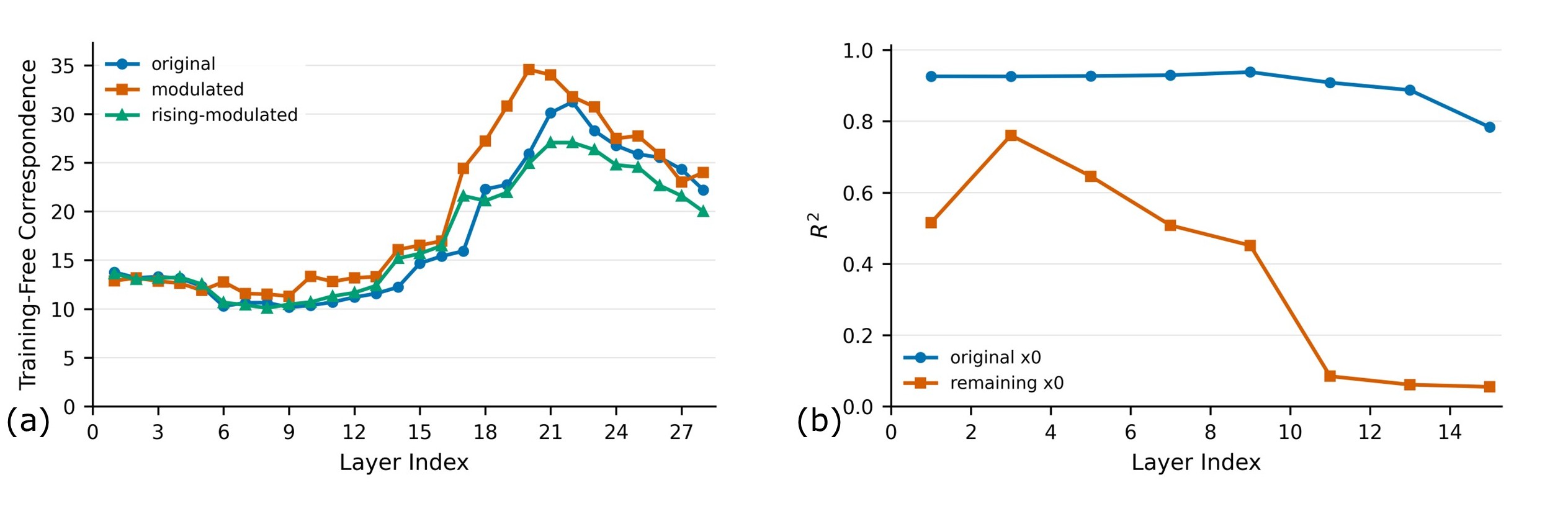}
    \caption{(a) Training-free correspondence probe results from the original and the modulated FM+$a45$ models.
    (b) Separability of FM+$a45$ model's MLP increments, measured using the remaining $x_0$ signal after deleting $x_0$ from the feature while minimizing the damage to $\epsilon$.}
    \label{fig:intervention-separability}
\end{figure}

Then, to validate that the prepared information is mainly consumed by the descent phase, we zero out MLP increments in the preparation phase and test both prediction loss and correspondence performance.
The prediction loss worsens (from 0.3852 to 0.4176), but the correspondence probe even improves (\Cref{fig:intervention-separability}(a)).
The interpretation of the latter is that the stored information effectively acts as a perturbation for the \textit{training-free} probe.
On the other hand, the worsening of the prediction loss demonstrates that the stored information is actively utilized during the descent phase.
For comparison, when we zero out the same number of layers in the rising phase instead, the loss becomes 0.4255, but the correspondence performance drops, indicating that it has a different mechanism.

With the role of the preparation phase clarified, we next provide a more robust way to measure the length of the preparation phase instead of visually observing the raw correspondence curve.
If we use the separability probe to separate $x_0$ and $\epsilon$ signals in MLP increments, we can observe from \Cref{fig:intervention-separability}(b) a generally descending trend of the remaining $R^2$ values after deletion.
We further define the last layer before the remaining $R^2$ drops below 0.1 as the final layer of the preparation phase.
This is a more reliable definition compared to the training-free correspondence probe curve, which is sensitive to post-hoc transforms such as whitening (\Cref{fig:separability-robustness} of Appendix).

\subsection{Rising Phase}
The key characteristic of the rising phase is the formation of the semantic abstract of the clean image, which we denote as $x_s$.
Note that $x_s$ is largely different from $x_0$, which is a VAE- or RGB-space signal.
Imagine doing nearest-neighbor correspondence on an RGB image directly.
The result will naturally be bad compared to doing it on a DINOv2~\citep{dino} feature.
This explains the difference between an abstract, conceptual, semantic $x_s$ and a VAE- or RGB-space $x_0$.

The behavior in this phase largely aligns with the current knowledge.
To provide additional information, we utilize the linear regression probe, where the source feature is the per-block features, and the target feature includes $x_0$, the peak-layer feature as a proxy of $x_s$, and DINOv2~\citep{dino} features as another external proxy of $x_s$.
We can observe from \Cref{fig:rising-descent}(a) that the $x_0$ curve constantly rises, while the $x_s$ curves (especially the DINOv2 curve) peak at roughly the same position as the correspondence peak.
This further strengthens what we have already known.

\subsection{Descent Phase}
The characteristic behavior of the descent phase is the drop of the correspondence performance.
The current interpretation is that the granularity of $x_s$ further becomes finer until the network exit.
However, our overall picture gives a different explanation: the drop is because the model writes the encoded prediction target into the residual, making the semantic $x_s$ harder to decode.

\begin{figure}[t]
    \centering
    \includegraphics[width=\linewidth]{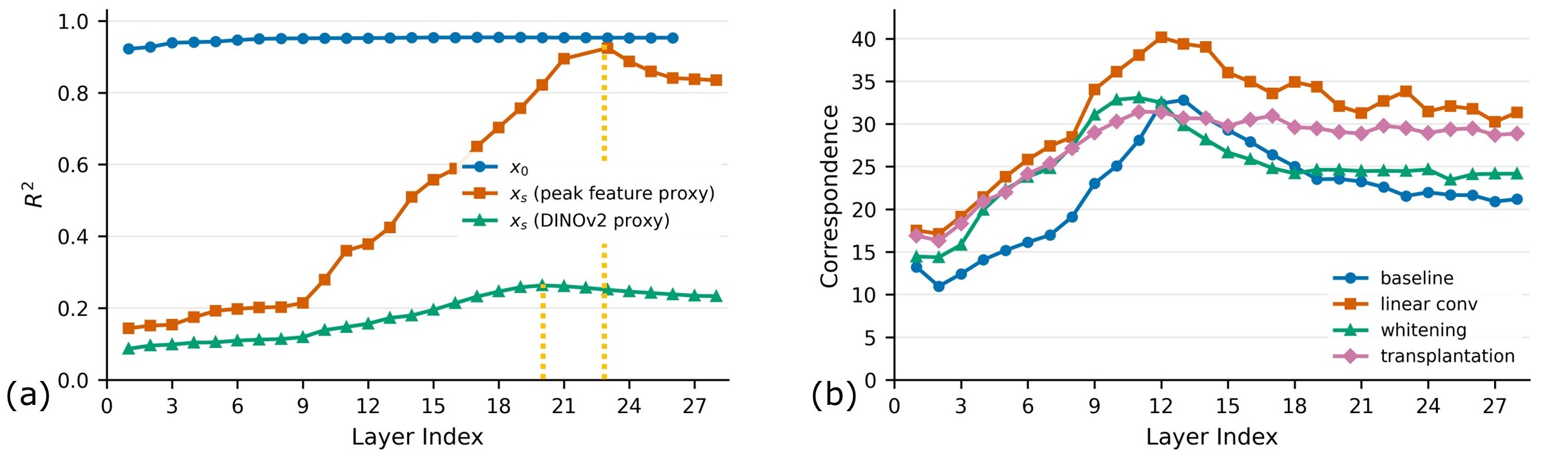}
    \caption{(a) Linear regression probe results of $x_0$, $x_s$ via peak feature proxy, and $x_s$ via DINOv2 proxy.
    (b) Correspondence probe results from the FM+$\epsilon$ (clean weighting) models, using the baseline training-free probe, linear convolution probe, whitening probe, and transplantation probe (projecting features into the peak feature's space).}
    \label{fig:rising-descent}
\end{figure}

To validate our explanation, we add a trainable convolutional layer to the correspondence probe, granting it linear decoding ability.
With results in \Cref{fig:rising-descent}(b), whilst the trainable layer brings overall improvement, the improvement is more significant in the descent phase, indicating a ``fixing'' effect introduced by the linear decoding ability.
A similar effect can be observed with whitening and transplantation (into the peak feature's linear space) probes.
In other words, the $x_s$ signal still exists to a large extent, only covered under more complex encoding.
Nevertheless, this behavior can coexist with the granularity change explanation.

\subsection{Effect of Prediction Target}
Next, we discuss how the overall picture predicts the potential impact of training setups on the depth profile.
The first prediction, related to prediction targets, is that a more hybrid target (such as $u=\epsilon-x_0$) leads to a longer preparation phase, forming a roughly symmetric pattern.

This prediction can be derived from two separate hypotheses.
(i) A prediction target that is closer to $\epsilon$ naturally requires more $\epsilon$ for the encoding, and vice versa, which further affects what should be prepared in the preparation phase.
Then, a hybrid prediction target requires more information to be prepared during the preparation phase, as it needs \textit{both} $\epsilon$ and $x_0$ for hints.
This gives the preparation phase a harder task, which results in a longer preparation phase.
(ii) The difference from the first hypothesis is that what needs to be prepared is determined by inference-time SNR.
For example, at a noisy SNR, the descent phase is able to infer $\epsilon$ directly from the residual, so it does not need to be prepared in advance.
On the other hand, the overall prediction task trivializes for purer targets at certain SNR, \textit{e.g.}, predicting $x_0$ will be too easy at a clean SNR to form meaningful optimization pressure, so the $x_0$-model's depth profile is mainly shaped by noisy SNR.
Since at noisy SNR, $\epsilon$ is not required to be prepared in advance, the $x_0$-model only learns to prepare $x_0$.
Then, a more hybrid prediction target creates more meaningful optimization pressure on both noisy and clean SNR, leading to longer preparation phases.

\begin{figure}[t]
    \centering
    \includegraphics[width=\linewidth]{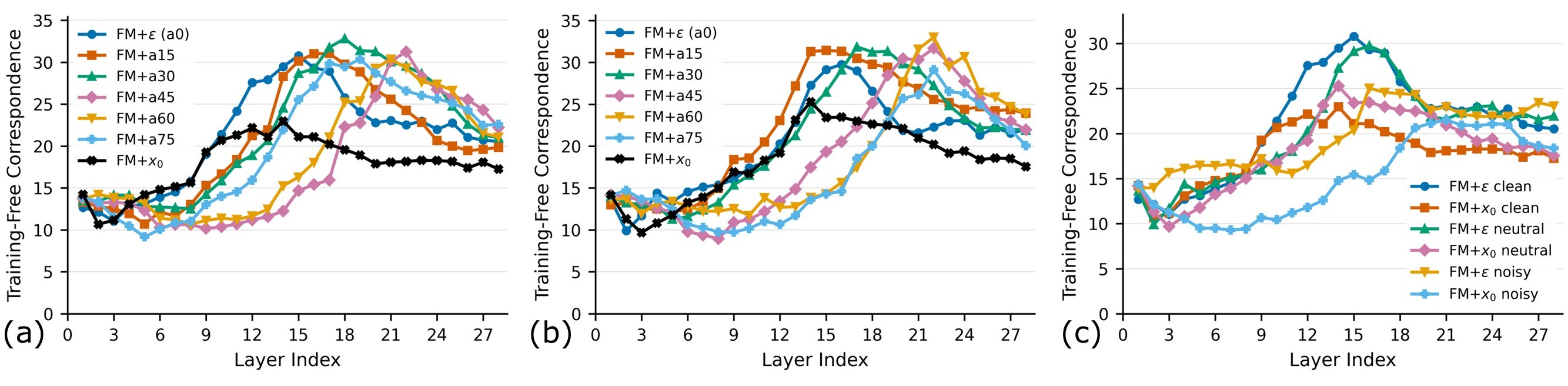}
    \caption{(a) Models trained using the basic min-SNR weighting (emphasizing cleaner samples) with different prediction targets.
    (b) Models trained using the neutral weighting with different prediction targets.
    (c) $x_0$- and $\epsilon$-models trained with three loss weighting methods.}
    \label{fig:target+weighting}
\end{figure}

First, we observe the predicted roughly symmetric pattern from two groups of models, one trained with the clean weighting (\Cref{fig:target+weighting}(a)) and the other with the neutral weighting (\Cref{fig:target+weighting}(b)).
We also complement the two figures with \Cref{fig:prediction-target-separability} in Appendix, which depicts the pattern using the separability probe.
Then, although the two hypotheses lead to the same result, their inner workings are different.
The first one suggests that the ``importance'' of $\epsilon$ and $x_0$ changes solely with the prediction target, while the second one predicts that the importance is more related to the SNR at which we do the inference.
Therefore, we make another causal intervention.
The target of intervention is still MLP increments in the preparation phase, but we do not zero out the entire increment.
Instead, we selectively replace the $\epsilon$ part with random noise while minimizing the damage to the $x_0$ part, or vice versa.
The result in \Cref{tab:target-loss-intervention} uses the same random seed across the same column for better alignment.
It shows a mixed pattern, suggesting the actual case might be a mixture of both hypotheses.

\begin{table}[t]
\centering
\caption{Causal intervention by replacing a certain component within the MLP increments with random noise.
The more severe loss increase is marked as \resultbest{bold}.
\textit{The same seed is shared across the same column} that controls both image sampling and noise sampling.
The absolute loss value is not comparable across models or SNR.
All interventions are done with FM+$x_0$ and FM+$\epsilon$ models trained under the neutral weighting.}
\label{tab:target-loss-intervention}
\begin{tabular}{@{}llllllllc@{}}
\toprule
\multirow{2}{*}{Setting} &
  \multicolumn{4}{c}{$log$SNR=1.69} &
  \multicolumn{4}{c}{$log$SNR=-1.69} \\
 &
  \multicolumn{1}{c}{seed 1} &
  \multicolumn{1}{c}{seed 2} &
  \multicolumn{1}{c}{seed 3} &
  \multicolumn{1}{c}{avg} &
  \multicolumn{1}{c}{seed 1} &
  \multicolumn{1}{c}{seed 2} &
  \multicolumn{1}{c}{seed 3} &
  avg \\ \midrule
$x_0$ (original) &
  0.0741 &
  0.0720 &
  0.0734 &
  0.0732 &
  0.3352 &
  0.3347 &
  0.3344 &
  0.3348 \\
$\epsilon$ (original) &
  0.3969 &
  0.3969 &
  0.3941 &
  0.3960 &
  0.0618 &
  0.0642 &
  0.0618 &
  0.0626 \\ \midrule
$x_0$ (replacing $x_0$) &
  0.1767 &
  \resultbest{0.1833} &
  0.1816 &
  0.1805 &
  \resultbest{0.8191} &
  \resultbest{0.8599} &
  \resultbest{0.8336} &
  \resultbest{0.8375} \\
$x_0$ (replacing $\epsilon$) &
  \resultbest{0.1838} &
  0.1813 &
  \resultbest{0.1841} &
  \resultbest{0.1831} &
  0.7542 &
  0.7758 &
  0.7828 &
  0.7709 \\ \midrule
$\epsilon$ (replacing $x_0$) &
  0.6127 &
  0.6162 &
  0.6274 &
  0.6188 &
  0.2258 &
  0.2323 &
  0.2314 &
  0.2298 \\
$\epsilon$ (replacing   $\epsilon$) &
  \resultbest{0.6245} &
  \resultbest{0.6277} &
  \resultbest{0.6305} &
  \resultbest{0.6276} &
  \resultbest{0.2350} &
  \resultbest{0.2360} &
  \resultbest{0.2441} &
  \resultbest{0.2384} \\ \bottomrule
\end{tabular}
\end{table}

\subsection{Effect of Loss Weighting}
The second prediction, related to loss weighting, is two-fold.
(i) A pure target might undergo certain degeneration under certain weighting.
(ii) A prediction target where $x_0$ and $\epsilon$ are not entirely symmetric will have the length of the preparation phase changed.

\textbf{Degeneration}:
Taking $\epsilon$ as an example of pure targets, it has a certain degeneration scenario.
At a very noisy SNR where the input is almost noise, the task of predicting the noise becomes trivial, as the model can almost do nothing and still get a low loss, thus creating weak optimization pressure.
When the loss weighting emphasizes an SNR section that corresponds to this degeneration scenario, the depth profile will become degenerated, as a good internal $x_s$ signal is unnecessary.
This prediction is validated by the observation in \Cref{fig:target+weighting}(c) that the $\epsilon$-model is normal at the neutral and the clean weighting, but degenerated at the noisy weighting.

\textbf{Length of Preparation Phase}:
When the loss weighting emphasizes a certain part of the SNR-axis, the task for the preparation phase may become harder.
For example, for a target that has more $\epsilon$, the task is \textit{harder} at a cleaner SNR, since there is \textit{less} amount of $\epsilon$ in the model input, which results in a \textit{longer} preparation phase.
This prediction is validated by the observation that the $a60$-model has a shorter preparation phase with an SNR-weighting that focuses on the clean part (\Cref{fig:target+weighting}(a)) than with a neutral weighting (\Cref{fig:target+weighting}(b)).
Different from the first effect that only applies to highly pure prediction targets, the second effect is more general.

\textbf{Inherent Asymmetry of $x_0$ and $\epsilon$}:
We can also observe from \Cref{fig:target+weighting} that there is a certain inherent asymmetry between $x_0$ and $\epsilon$.
For instance, even at a neutral weighting, $\epsilon$ is normal while $x_0$ is degenerated, and $a30$ is not symmetric to $a60$.
It is still unclear why such an asymmetry exists, but we have confirmed that the VAE scaling factor mismatch issue is \textit{not} the cause.
For more details, refer to \Cref{sec:app-vae-scaling}.

%% file: sections/6_example.tex
\section{Practical Corollaries and Illustrative Applications}

\begin{table}[t]
\centering
\caption{Generation performance of aligning REPA at different layers during training. Best marked as \resultbest{bold} and second marked as \resultsecond{underlined}. The models are trained for 100k steps, and the evaluation is done on 10k samples generated using CFG=1.5.}
\label{tab:repa}
\begin{tabular}{@{}lccccccc@{}}
\toprule
REPA Layer & 5      & 7     & 9      & 11     & 13          & Dynamic & Inverted Dynamic  \\ \midrule
FID$\downarrow$                  & 15.56  & 13.4  & 12.49  & \resultsecond{12.31}  & 12.33 & \resultbest{11.43} & 12.33 \\
IS$\uparrow$                   & 101.96 & 116.6 & 124.56 & \resultsecond{125.63} & 123.6 & \resultbest{133.2} & 124.34 \\ \bottomrule
\end{tabular}
\end{table}

REPA proposes to introduce an auxiliary loss that explicitly aligns diffusion features to external self-supervised features, such as DINOv2~\citep{dino} features.
The optimal placement of the REPA loss can be derived from our findings.
Specifically, we notice the reported optimal layer in the original paper~\citep{REPA} is around the interface between the preparation phase and the rising phase.
To test our mechanistic findings in a real application scenario, we measure the length of the preparation phase of the $a60$-model trained with the basic min-SNR weighting.
The length is 9 at 200k training steps, so we put the REPA loss at the last layer of the preparation phase.
The results in \Cref{tab:repa} show that our prediction is around the optimal placement but not the exact optimum.

Additionally, we have also noticed that the length of the preparation phase gradually shortens throughout the training (\Cref{fig:prep-len-steps} in Appendix).
So a more aggressive prediction of our findings is that we can dynamically move the REPA placement to earlier layers as the training goes on.
For the concrete $a60$-model, we adopt a setting of \{13@0$\rightarrow$11@30k$\rightarrow$10@40k$\rightarrow$9@50k\} where the REPA loss stays at layer 9 after 50k steps.
This choice is based on the measured preparation phase length at different training steps rather than being tuned.
The \textit{Dynamic} column in \Cref{tab:repa} achieves 11.43 FID score and 133.2 IS score, surpassing the previous optimum and validating the prediction.
This can also explain why our previous prediction of layer 9 is not the exact optimum: in earlier training steps, a deeper layer would be better.
If we use a static alignment layer, it might be better to take the average across the entire training process.
In addition, we also try the inverted dynamic alignment, which is significantly worse than the proposed dynamic alignment, showing that the benefit is from a correct schedule rather than being dynamic itself.
As for why REPA can work, our findings may provide some new angles (\Cref{fig:repa-probe} in Appendix), but we leave that for future work.

%% file: sections/7_conclusion.tex
\section{Conclusion \& Limitation}
This work studies the depth profile of diffusion backbones.
We provide new empirical results that cover how the depth profile changes with training setups.
Furthermore, we offer novel and more complete mechanism findings, validated by the new observations.
The new findings may potentially impact applications that rely on this model property, such as the layer selection of REPA.

Note that our work is based on DiT/SiT-style backbone architectures, which creates a gap between our work and larger-scale models such as Flux or Qwen-Image.
Additionally, although we have tried to employ more diverse probes to get a more complete picture, these probes are still biased in nature, as all model probing studies are.
Overall, our findings are falsifiable as the scope expands.

%% file: sections/appendix.tex
\section{Fixing VAE Scaling Factor}
\label{sec:app-vae-scaling}

To expose the misalignment between $x_0$ and $\epsilon$ brought by the VAE scaling factor 0.18215, we use linear regression to measure $x_t$, which is not even sent into the model yet.
The result shows that the $x_0$ signal is inherently weaker ($R^2=0.4451$) compared to the $\epsilon$ signal ($R^2=0.6052$).
This is because the common practice multiplies the VAE-compressed $x_0$ by 0.18215, which is used by the original Stable Diffusion~\citep{rombach2022high} team to align the variance of $x_0$ and $\epsilon$ and \textit{only applies to their data}.
The currently common practice of using the same number is thus in fact improper and will lead to variance misalignment.
Nevertheless, the impact on generation performance is often relatively negligible, which might explain why the specific number is used so widely, despite the fact that it is derived from a very specific case.

\begin{figure}[t]
    \centering
    \includegraphics[width=\linewidth]{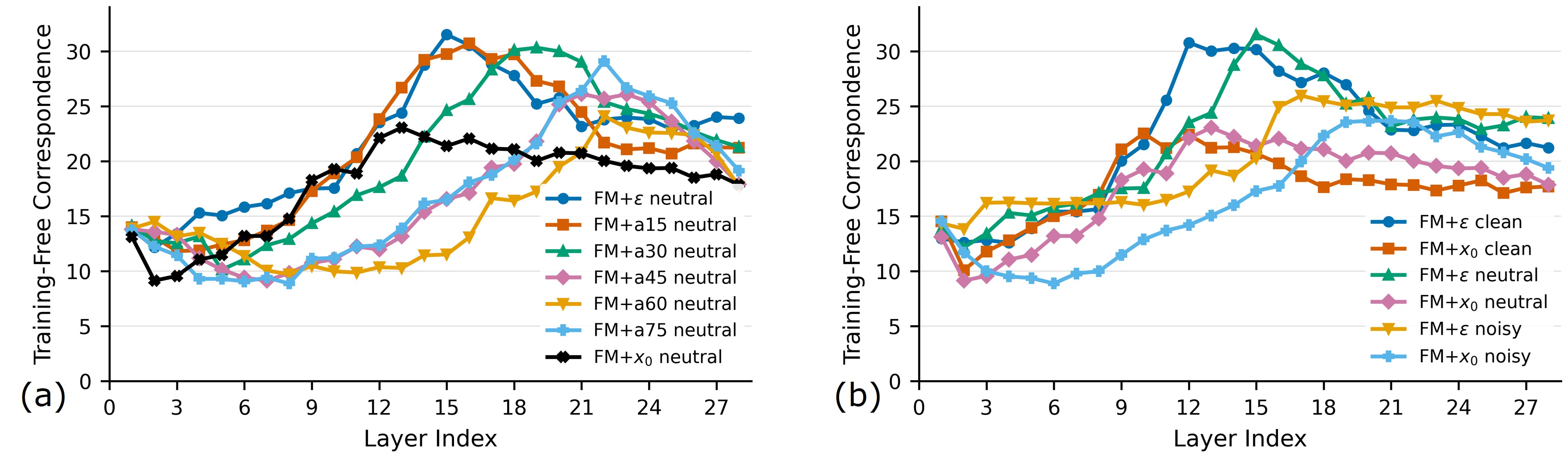}
    \caption{(a) Models of different prediction targets trained with a fixed VAE scaling factor, under the neutral loss weighting.
    (b) Models trained under different loss weighting with a fixed VAE scaling factor.}
    \label{fig:vae-fix}
\end{figure}

\begin{figure}[t]
    \centering
    \includegraphics[width=\linewidth]{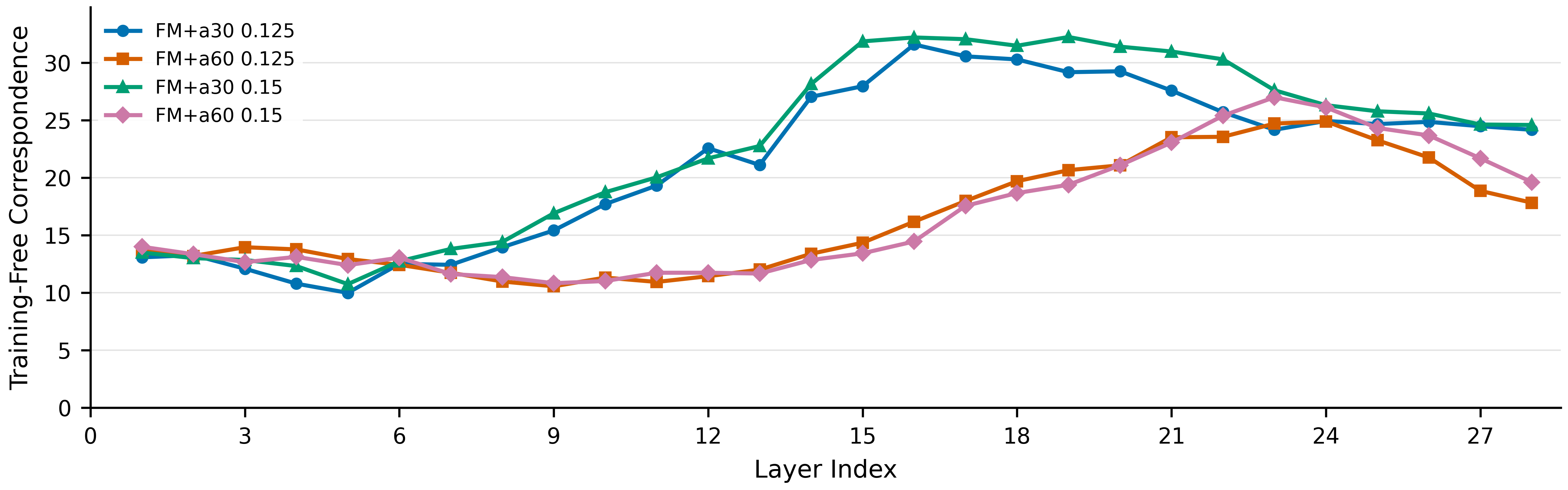}
    \caption{Moving VAE scaling factor to the opposite direction, making it smaller than 0.18215 (0.15 or 0.125). This does not fix the asymmetry between $x_0$ and $\epsilon$ either.}
    \label{fig:vae-opposite}
\end{figure}

To fix this effect, we have trained models with a properly designed scaling factor (0.218265 for ImageNet~\citep{imagenet}).
The new results include a prediction target comparison and a loss weighting comparison.
The results in \Cref{fig:vae-fix} show that fixing the scaling factor mismatch does not remove the asymmetry between $x_0$ and $\epsilon$ targets:
(i) In \Cref{fig:vae-fix}(a), $a30$- and $a60$-models are still asymmetric.
(ii) In \Cref{fig:vae-fix}(b), the $x_0$-model with the neutral weighting still suffers from degeneration.
We also explore moving the VAE scaling factor to the opposite direction, assuming that there is indeed some inherent asymmetry, but it can be fixed by a certain VAE scaling factor.
The corresponding results in \Cref{fig:vae-opposite} demonstrate that this modification does not fix the asymmetry either.
Hence, it is still unclear why there is an inherent asymmetry between $x_0$ and $\epsilon$, but at least we are sure moving along the VAE scaling factor axis cannot reach the root cause.
Revealing the secret might unlock deeper understanding of diffusion training dynamics, so this would be an interesting future topic.
Nevertheless, we notice that fixing the VAE scaling factor does improve generation, so we still want to highlight the importance of using a VAE scaling factor that matches the specific training data.

\section{Additional Probing Results}
\label{sec:app-probe-results}

\begin{figure}[t]
    \centering
    \includegraphics[width=\linewidth]{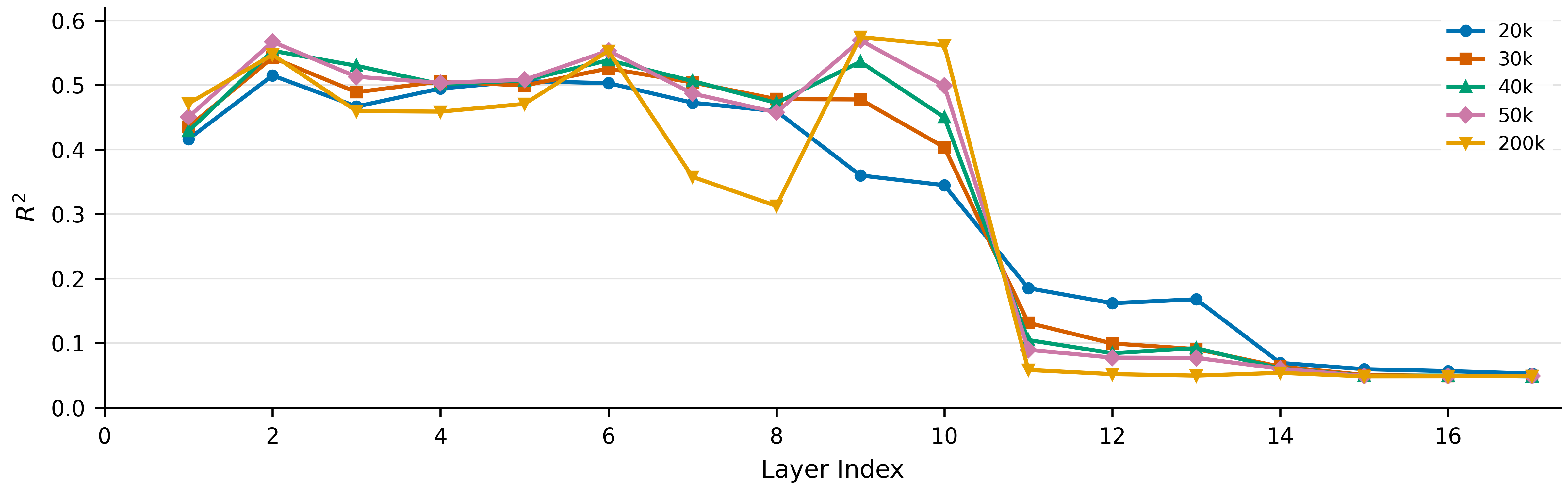}
    \caption{Using separability probe to measure the length of the preparation phase as the training goes on. We can observe that the preparation phase gradually shortens.}
    \label{fig:prep-len-steps}
\end{figure}

We use separability probe to measure how the length of the preparation phase changes with the training steps.
The results in \Cref{fig:prep-len-steps} suggest that the preparation phase gradually shortens as the training goes on, but the shortening rate slows down.
This result also serves as evidence for our dynamic REPA alignment design.

\begin{figure}[t]
    \centering
    \includegraphics[width=\linewidth]{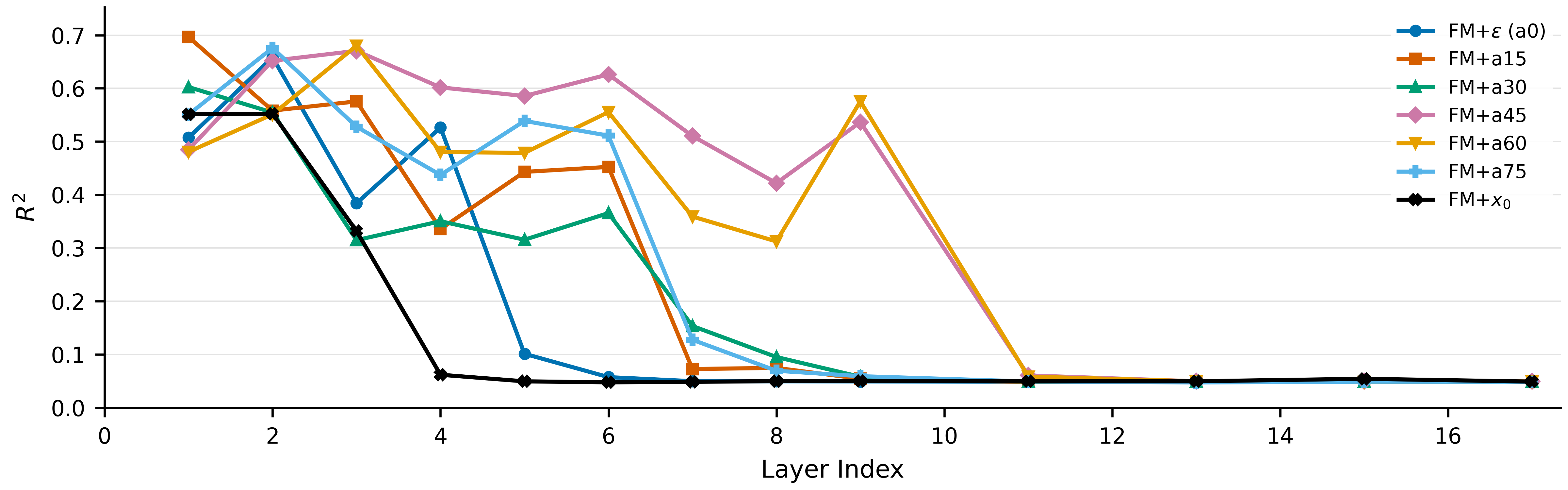}
    \caption{Using separability probe to depict the roughly symmetric pattern caused by prediction targets. The models are trained with the clean weighting.}
    \label{fig:prediction-target-separability}
\end{figure}

\Cref{fig:prediction-target-separability} serves as complementary information to \Cref{fig:target+weighting}(a,b) by showing the roughly symmetric patterns via separability probing.

\begin{figure}[t]
    \centering
    \includegraphics[width=\linewidth]{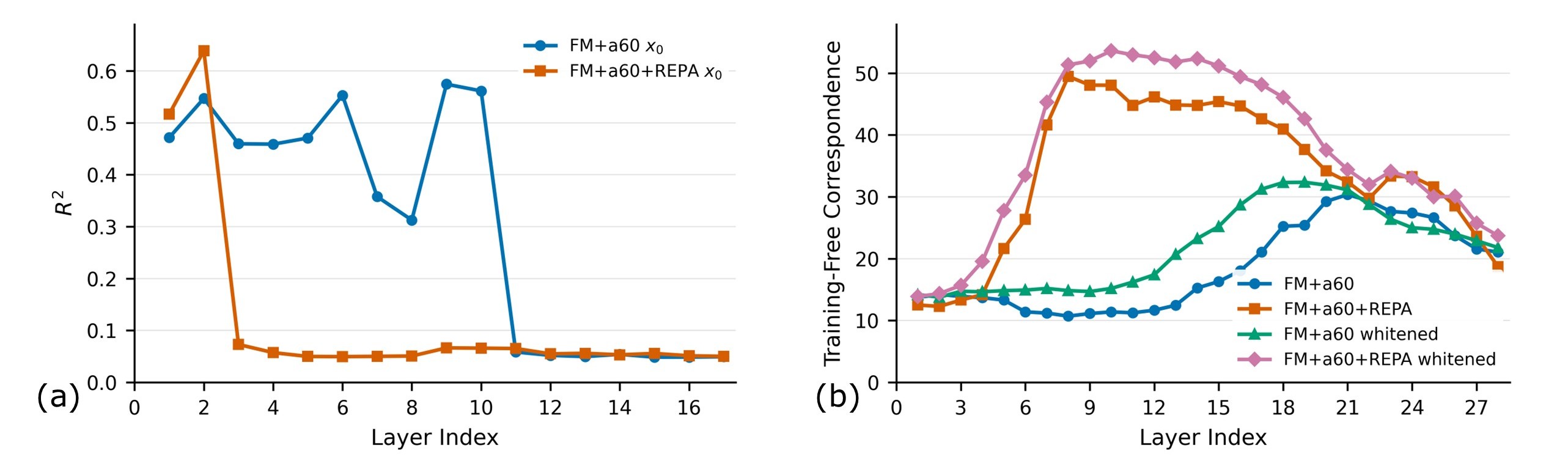}
    \caption{(a) Using separability probe to measure the REPA-augmented model, finding that the separability in early layers significantly improves.
    (b) The correspondence performance as well as the curve shape are also changed greatly.}
    \label{fig:repa-probe}
\end{figure}

To provide preliminary empirical observations for future work on why REPA can work, we use both separability probe and the training-free correspondence probe to measure the $a60$ model trained with the dynamic REPA.
It can be observed from \Cref{fig:repa-probe} that the separability in early layers improves significantly, and the downstream task performance is also greatly boosted.
Hence, separability between $x_0$ and $\epsilon$ in early-layer MLP increments might serve as a way to reveal a new explanation for why REPA can work.

\begin{figure}[t]
    \centering
    \includegraphics[width=\linewidth]{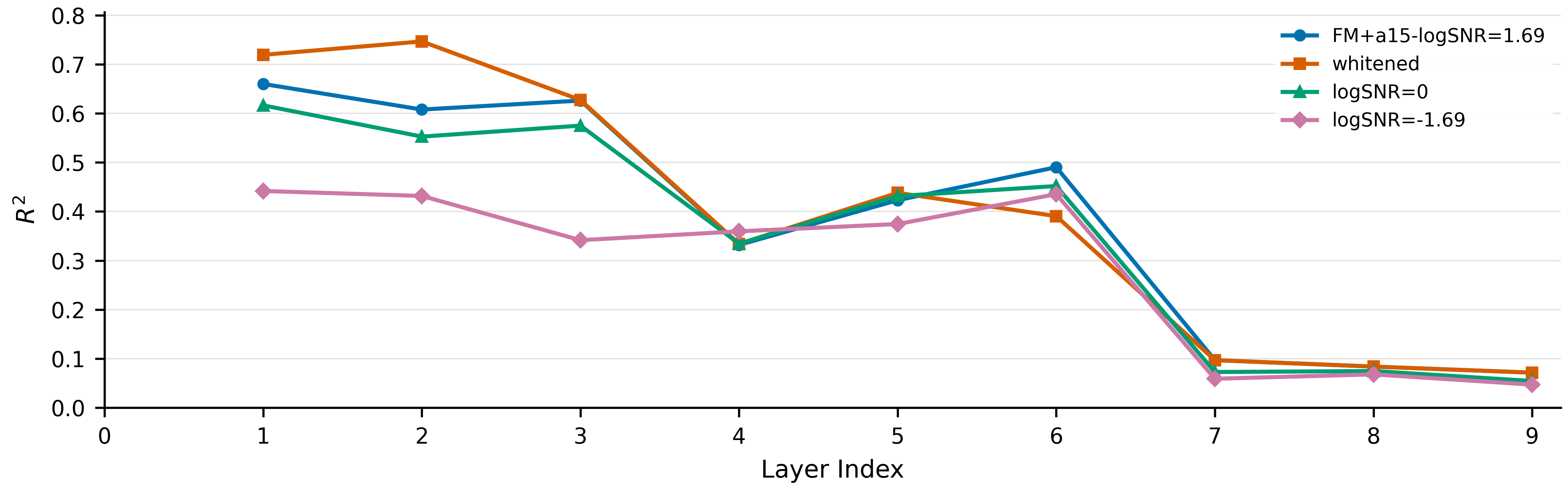}
    \caption{Measuring the separability of the same FM+$a15$ model with whitening and with different SNR. This result shows the robustness of separability probe as the definition of the length of the preparation phase.}
    \label{fig:separability-robustness}
\end{figure}

To show that the separability probe dropping below a certain threshold is a robust definition for the length of the preparation phase, we measure the separability of the same model with whitening and with different SNR.
The results in \Cref{fig:separability-robustness} clearly demonstrate the robustness.

\begin{table}[t]
\centering
\caption{We measure the generation performance of well-conditioned and degenerated models. The results appear to suggest that degeneration does not necessarily lead to poor generation.
All models use FM paths.}
\label{tab:generation}
\begin{tabular}{@{}lcccccc@{}}
\toprule
Model & $x_0$-clean & $x_0$-neutral & $x_0$-noisy & $\epsilon$-clean & $\epsilon$-neutral & $\epsilon$-noisy \\ \midrule
FID$\downarrow$   & 62.66   & 60.55      & 72.16      & 61.62    & 63.22       & 82.71       \\
IS$\uparrow$    & 50.62   & 59.61      & 41.16      & 55.19    & 52.81       & 31.80       \\ \bottomrule
\end{tabular}
\end{table}

We have shown that certain loss weighting can cause degeneration.
To know whether degeneration will also make generation worse, we evaluate some models with generation metrics.
The results in \Cref{tab:generation} suggest that degeneration does not necessarily lead to poor generation, or its impact is relatively small.
Hence, if diffusion features are not of interest, degeneration caused by certain training setups is acceptable to some extent.

We have also uploaded additional probing results in .csv format in the supplementary material.

\section{Implementation Details}
\subsection{Model \& Training}

\textbf{Model Architecture}.
One notable implementation detail is how the timestep input $t$ is embedded into the network.
We first cast the original $t$ into normalized $log$SNR and embed it instead.
This effectively aligns the two sampling paths, which originally use different timestep definitions.

\textbf{Training}.
The training uses ImageNet~\citep{imagenet} at 256$\times$256 resolution as the training set.
We use the same VAE as used in DiT~\citep{Peebles23dit} and SiT~\citep{sit}.
We use AdamW~\citep{kingma2014adam} optimizer, with a constant 1e-4 learning rate, betas as (0.9, 0.999), eps as 1e-8, wd as 0.
We add a learning rate scheduler that linearly warms up through the first 5000 steps and then stays constant afterward.
The batch size is 256.
Additionally, we implement gradient clipping with global norm set to 1.0.
The $log$SNR sampling span is [-10.12, 9.21], which is calculated from the original DDPM~\citep{ho2020denoising} setting.

\subsection{Probes}
\textbf{Training-Free \& Convolutional Correspondence}.
Training-free semantic correspondence is evaluated on the SPair-71k~\citep{min2019spair} dataset using nearest neighbor matching via cosine similarity.
The results are reported under PCK@0.1, which counts the percentage of correctly predicted samples, where a prediction is considered correct if it lies within a radius of the ground truth point based on 0.1 times the image width and height.
For the convolutional correspondence probe, we add one linear convolutional layer before nearest neighbor to refine the features, which is trained for 500 steps only.

\textbf{Whitening}.
Given a batch of features $f$, we first accumulate the mean $\mu$ and centered covariance $C$.
$C$ is factorized as $C=V\Lambda V^T, \Lambda=\text{diag}(\lambda_1, \cdots,\lambda_d)$.
Then, we calculate the whitening matrix as $M=V\text{diag}(\lambda_1+\kappa\lambda_{\text{max}},\cdots,\lambda_n+\kappa\lambda_{\text{max}})^{-1/2}V^T$, where $\kappa$ is a relative truncation threshold.
Finally, the per-layer feature is processed as $f'=(f-\mu)M$.

\textbf{Transplantation}.
For a source feature $f_s$ and a reference feature $f_r$, we transplant the source feature into the linear space of the reference feature.
First, accumulate the cross-covariance between the two features.
Then obtain a projection matrix using a closed-form least-squares problem.

\textbf{Linear Regression}.
For this probe and the following probes, the experiments are done on ImageNet~\citep{imagenet}.
The diffusion features within DiT backbones are a token sequence, \textit{i.e.}, a sequence of vectors in the shape of $\mathbb{R}^{c\times hw}$.
We first align the size of source and target sequences using bilinear interpolation.
Then, on the two aligned features, each token in the sequence serves as a linear regression target.
After regression, the reported $R^2$ is computed separately for each channel over all tokens and then averaged.



\textbf{Separability Probe}.
Denote the per-layer feature as $f_s$, the feature to delete as $f_t$, and the feature to protect as $f_p$.
First, compute the standardized cross-covariance between $f_s$ and $f_t$, and take an orthonormal basis of its column space as $S_t$.
Then, $f_s$ is orthogonally projected to remove this subspace:
\begin{equation}
    \label{equ:separability}
    f_s'=f_s(I-S_tS_t^T)
\end{equation}

\textbf{Geometry Probe}.
Given a feature and another reference feature, the geometry probe measures the relative norm and the angle between the given feature and the reference feature.
The norm is calculated as follows: first calculate the norm of each token in the sequence, and then take the average over the sequence.
The angle is calculated as follows: we first compute the centered covariance GRAM of the entire token sequence, and then use the direction of its top eigenvector as the direction of this whole feature. Afterward, we compute the cosine angle between the two given features.